\documentclass[runningheads]{llncs}

\usepackage{eccv}

\usepackage{eccvabbrv}

\usepackage{graphicx}
\usepackage{booktabs}

\usepackage{hyperref}

\usepackage{orcidlink}

\usepackage{multirow}
\usepackage{floatrow}

\newfloatcommand{capbtabbox}{table}[][\FBwidth]

\usepackage{subcaption}

\makeatletter
\newcommand{\thickhline}{%
    \noalign {\ifnum 0=`}\fi \hrule height 1pt
    \futurelet \reserved@a \@xhline
}

\makeatother

\begin{document}

\title{Decomposition Betters Tracking Everything Everywhere} 

\titlerunning{Decomposition Betters Tracking Everything Everywhere}

\author{Rui Li\orcidlink{0009-0008-3102-0932} ~~~~~~~~~ Dong Liu\orcidlink{0000-0001-9100-2906}}

\authorrunning{Rui Li et al.}

\institute{University of Science and Technology of China, Hefei, China \\
\email{liruid@mail.ustc.edu.cn,dongeliu@ustc.edu.cn} \\
\url{https://github.com/qianduoduolr/DecoMotion}}

\maketitle

\begin{abstract}
Recent studies on motion estimation have advocated an optimized motion representation that is globally consistent across the entire video, preferably for every pixel. This is challenging as a uniform representation may not account for the complex and diverse motion and appearance of natural videos. We address this problem and propose a new test-time optimization method, named DecoMotion, for estimating per-pixel and long-range motion. DecoMotion explicitly decomposes video content into static scenes and dynamic objects, either of which uses a quasi-3D canonical volume to represent. DecoMotion separately coordinates the transformations between local and canonical spaces, facilitating an affine transformation for the static scene that corresponds to camera motion. For the dynamic volume, DecoMotion leverages discriminative and temporally consistent features to rectify the non-rigid transformation. The two volumes are finally fused to fully represent motion and appearance. This divide-and-conquer strategy leads to more robust tracking through occlusions and deformations and meanwhile obtains decomposed appearances. We conduct evaluations on the TAP-Vid benchmark. The results demonstrate our method boosts the point-tracking accuracy by a large margin and performs on par with some state-of-the-art dedicated point-tracking solutions. 

\end{abstract}

\section{Introduction}
\label{sec:intro}
Video content comprises two primary components: dynamic and static scenes. In dynamic scenes (also denoted as dynamic objects), objects exhibit movement or change over time, and the motion between video frames results from a combination of camera movement and the object’s inherent motion. Conversely, in static scenes, objects remain unchanged, and inter-frame motion is solely influenced by the camera’s rigid movement. Compared to static scenes, dynamic scenes are more intricate in terms of both motion and appearance.

However, the advanced studies for establishing point correspondences in the last few years have always dealt with different motions in two scenes with a unified framework. For example, feature matching methods\cite{wang2019learning,jabri2020space,xu2021rethinking} try to design learning methods for the dense features. Despite achieving good performance on tasks involving dynamic object tracking, these methods struggle to model consistent motion in static scenes. Dense optical flow estimation methods regard pixel correspondences as a regression problem, and usually train a convolutional neural network on a rendered video dataset\cite{dosovitskiy2015flownet}, which are further improved by constructing a pyramid cost volume\cite{sun2018pwc}, iterative inference\cite{teed2020raft}, and multi-frame prediction\cite{shi2023videoflow,harley2022particle}. However, these methods still show unsatisfactory robustness in real-world videos, especially on dynamic objects.

In a recent study, OmniMotion \cite{wang2023omnimotion} proposes a novel test-time optimization method, which utilizes reliable correspondences and RGB frames as supervision to learn a globally consistent neural radiance field~\cite{mildenhall2021nerf} with learned transformation. The representation stores entire color and quasi-3D geometry information in canonical space, along with transformation can establish long-range pixel trajectories even facing occlusions. While promising, the method is optimized with a unified neural field and transformation, which results in sub-optimal performance of estimating the highly non-rigid motion of moving objects.

In this work, intending to learn accurate point correspondences, we propose a decoupled representation that divides static scenes and dynamic objects in terms of motion and appearance. Leveraging the significant difference between them, we explicitly build individual 3D neural radiance fields for representation, and we carefully design the transformation and representation functions for each scene. More specifically, for the dynamic objects, taking into account the complex inter-frame motion and dramatic appearance changes, we utilize more non-linear layers to approximate the non-rigid motion. Besides, we additionally encode the features in the field for better representation, along with a feature rendering constraint to alleviate the problem of imprecise correspondences on dynamic objects. For the static scenes, we approach the rigid-motion by a simpler network with affine transformation and model the confidence of whether the 3D points are changing or moving. With this confidence, we further fuse two neural radiance fields and transformations to obtain the final appearance and motion representations. Such a design enables DecoMotion to establish the trajectory of any pixel within the whole video more accurately, especially on dynamic objects. Additionally, the observed appearance decomposition enables its application to video tasks like video inpainting, and object removal.

In summary, the main contribution of this work lies in: (i) We propose an optimized 3D video representation that decomposes video into dynamic objects and static scenes for better tracking any pixel; (ii) For dynamic objects, we propose to encode and render discriminative and temporal consistent features to rectify the non-rigid transformation; (iii) We demonstrate our method quantitatively on the TAP-Vid benchmark \cite{doersch2022tap}. Our method significantly enhances point-tracking accuracy and shows competitive performance.

\section{Related work}

\noindent\textbf{Optimization-based motion estimation.} 
In the early days, motion estimation results are mostly obtained by optimizing the pairwise non-parametric constraint equations \cite{baker2004lucas,bouguet2001pyramidal,horn1981determining}, which are improved by optimizing motion globally over an entire video. Particle video \cite{sand2008particle} generates a collection of semi-dense long-range trajectories based on initial optical flows, while ParticleSfM \cite{zhao2022particlesfm} optimizes long-range rigid-motion derived from pairwise optical flows within a Structure-from-Motion (SfM) framework. By optimizing a globally consistent canonical volume and coordinate transformation with reliable optical flows, OmniMotion \cite{wang2023omnimotion} tries to deal with the non-rigid motion of dynamic objects. However, the results obtained by this method are not yet satisfactory.

\noindent\textbf{Matching-based motion estimation.} 
By matching between video frames, there are two dominant approaches: feature matching~\cite{wang2019learning,jabri2020space,vondrick2018tracking,li2019joint,li2023spatial} and dense optical flow~\cite{dosovitskiy2015flownet, ilg2017flownet, sun2018pwc,teed2020raft,xu2022gmflow}. In feature matching, for example,  by designing the pretext tasks like frame reconstruction~\cite{vondrick2018tracking,lai2020mast}, cycle-consistent tracking~\cite{wang2019learning,jabri2020space}, and contrastive learning~\cite{xu2021rethinking}, dense representations are learned for matching in a self-supervised manner. On the other hand, optical flow estimators regard it as a regression problem, and employ the cost volumes to find pixel matching.  FlowNet \cite{dosovitskiy2015flownet} first leverages a rendered video dataset to train a regression model, which further encourages the following methods to improve it by building pyramid cost volume\cite{sun2018pwc}, iterative inference\cite{teed2020raft}, and multi-frame prediction\cite{shi2023videoflow,harley2022particle,zheng2023pointodyssey,bian2023context}.

\noindent\textbf{Dynamic neural representation.} 
Our method shares similar spirits with recent studies about dynamic neural representation using coordinate-based neural networks \cite{park2021nerfies,li2021neural,li2023dynibar}.  For instance, Nerfies \cite{park2021nerfies} augments NeRF \cite{mildenhall2021nerf} by optimizing a time-dependent continuous volumetric deformation field that warps each observed point into a canonical 5D NeRF. NSFF \cite{li2021neural} proposes to represent a dynamic scene by combining static NeRF with scene flow-based dynamic NeRF, and is further improved by adopting a volumetric image-based rendering framework \cite{li2023dynibar}. However, the studies in this direction are mostly designed for problems such as space-time novel view synthesis, while our method aims to establish long-range accurate motion trajectories.

\section{Method}

This work presents a decoupled motion representation, we name it as DecoMotion~(Decomposing Motion). As follows, we will first revisit the OmniMotion \cite{wang2023omnimotion}, and then further introduce the proposed decoupled representation. The specific optimization process will also be presented in the next sections.


\subsection{Representation in OmniMotion}

OminiMotion \cite{wang2023omnimotion} proposes a quasi-3D global motion representation, i.e., a data structure that can encode not only the appearance but also the inter-frame motion of all points in the scene. Such a representation can provide accurate and consistent motion trajectories even when the query pixels are occluded.
\subsubsection{Canonical 3D volume.} OmniMotion represents the scenes with a canonical 3D volume $G$. Following  NeRF~\cite{mildenhall2021nerf}~, OmniMotion defines a coordinate-based network $F_{\theta}$ that maps any 3D point $u \in G$ to the corresponding color $c$ and density value $\sigma$. The density indicates the quasi-3D scene geometry of the video.

\subsubsection{Transformation.}
Meanwhile, OmniMotion defines a continuous invertible transformation $\mathcal{T}_i$ that maps the local 3D points $x_i$ in the local volume $L_i$ to the canonical space: $u = \mathcal{T}_{i}(x_i)$.  The $u$ is a time-independent 3D coordinate and can be considered as a globally consistent index of a particular point in the scene. By combining these invertible mappings, a 3D point can be mapped from one local volume frame $L_i$ to another local volume $L_j$, i.e:

\begin{equation}
  x_j=\mathcal{T}_j^{-1} \circ \mathcal{T}_i\left(x_i\right).
  \end{equation}

\subsubsection{Motion rendering.}
Given a global representation $G$ of the video and a transformation $\mathcal{T}$, OminiMotion predicts the motion by volume rendering. Specifically, for the query pixel $p_i$,  a set of local 3D points $\{x_i^k\}_{k=1}^K$ is sampled along the ray by a fixed, orthographic camera \cite{wang2023omnimotion}. These points are further mapped to the local volume $L_j$ using the invertible projections $\mathcal{T}_i$ and $\mathcal{T}_j$. These mapped 3D points $\{x_j^k\}_{k=1}^K$ are finally aggregated with alpha compositing:


  \begin{equation}
    \hat{\boldsymbol{x}}_j=\sum_{k=1}^K T_k \alpha(\sigma_k) x_j^k, \text { where } T_k=\prod_{l=1}^{k-1}\left(1-\alpha(\sigma_l)\right).
    \label{eq:alpha_comp}
    \end{equation}

The $\alpha$ value for $k$-th sample as $\alpha(\sigma_k)=1-\exp \left(-\sigma_k\right)$. The density and color are queryed as: $(\sigma_k,c_k) = F_{\theta}(\mathcal{T}_i(x_i^k))$, we omit $i$ when there is no ambiguity. To get the final motion estimation result $\hat{p}_j$, $\hat{\boldsymbol{x}}_j$ is directly projected back to the 2D image plane. In the same way, the image space color $\hat{C}_i$ for $p_i$ can be obtained by compositing $c_k$.
The representation uses reliable correspondences as well as the original RGB frames as labels for optimization. Given any novel coordinate in the frame, the model can render consistent full-length motion in the video.

\begin{figure}[!t]
  \centering
  \includegraphics[width=0.95\textwidth]{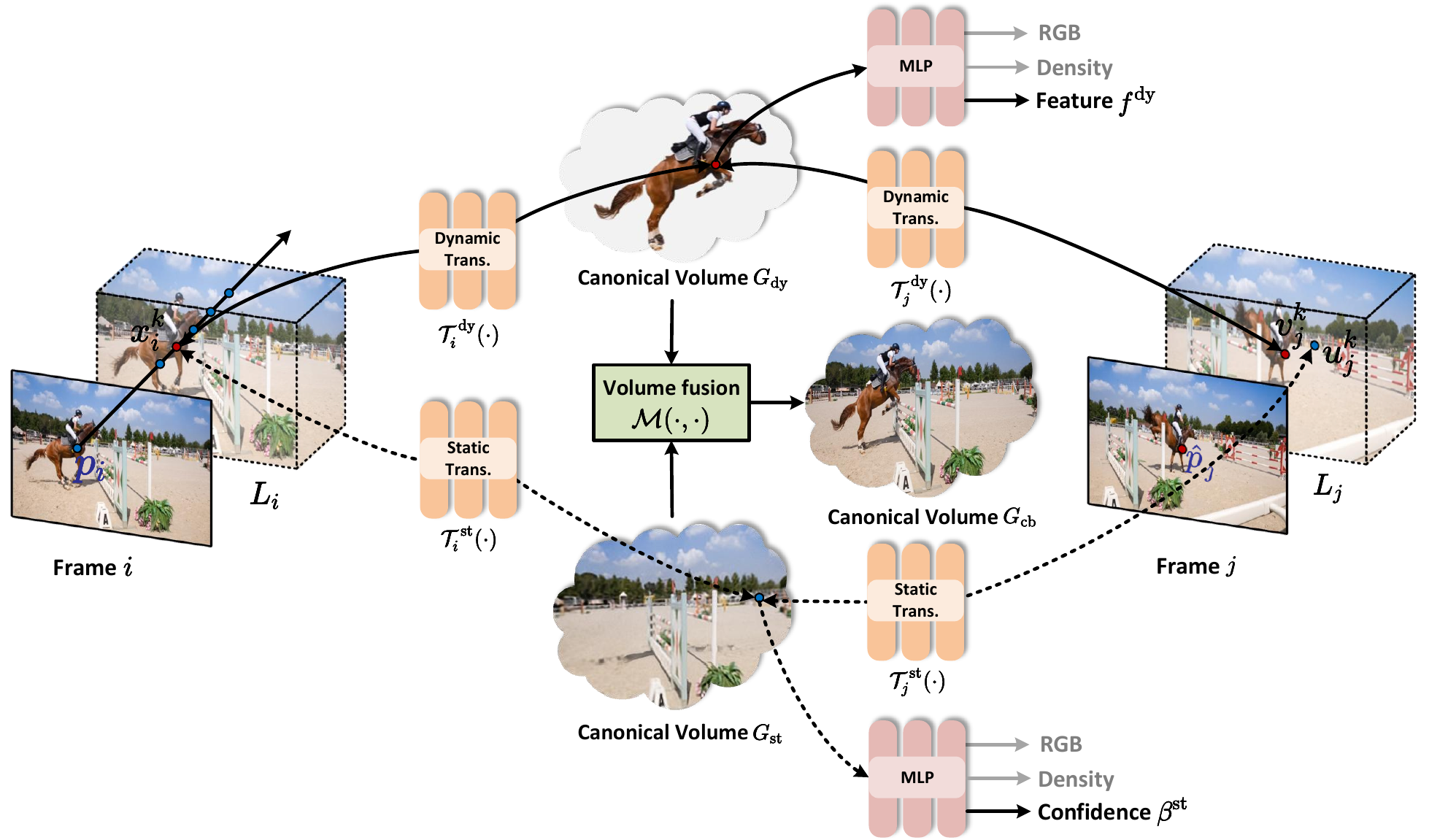}
  \caption{Overview of the proposed decoupled representation. We explicitly define two separate 3D canonical volumes $G_{\text{dy}}$ and $G_{\text{st}}$ to respectively characterize dynamic objects and static scenes in videos. In each representation, in addition to color and density, the $G_{\text{dy}}$ encodes the feature $f^{\text{dy}}$ to better represent dynamic objects, and feature rendering loss is further proposed to rectify the non-rigid transformation. The $G_{\text{st}}$ stores the $\beta^{st}$ determining the confidence of being a static point. In each canonical volume, we carefully design transformation functions $\mathcal{T}^{\text{dy}}$ (solid line) and $\mathcal{T}^{\text{st}}$ (dash line) to map each local 3D point $x_i^k$ along the ray of the point $p_i$ to $u_j^k, v_j^k$ in another local 3D volume $L_j$. In order to render the 2D correspondence $\hat{p}_j$ for $p_i$, we get the canonical volume $G_{\mathrm{cb}}$ of final representation by volume fusion with $\beta^{st}$. Set of 3D points $\{u_j^k\}_{k=1}^K,\{v_j^k\}_{k=1}^K$ mapped from $\{x_i^k\}_{k=1}^K$ are aggregated by alpha compositing in the $G_{\mathrm{cb}}$ (see Eq.(\ref{eq:ax})), and are projected to the image plane.}
  \label{fig:framework}
\end{figure}

\subsection{Decoupled representation}
As analyzed above, dynamic objects and static scenes are different in terms of movement and appearance. Therefore, we explicitly decouple the video into two parts, and optimize representations for static scenes and dynamic objects separately. The final representation is obtained by volumetric fusion. The overview of our method is shown in Figure \ref{fig:framework}.


\subsubsection{3D volume of static scenes.}
In contrast to a single global 3D volume, we maintain canonical volume $G_{\text{st}},G_{\text{dy}}$ for static scenes and dynamic objects respectively. For static scenes, $G_{\text{st}}$ refers to the design of ~\cite{wang2023omnimotion}, i.e., accessing the color  $c^{\text{st}}$ and the density $\sigma^{\text{st}}$. At the same time, $G_{\text{st}}$ also stores the confidence $\beta^{\text{st}}$ of whether each 3D point in the canonical volume is a static point or not through network $F_\theta^{\text{st}}$. The pixel color $\hat{C}^{\text{st}}_i$, motion mask $\hat{B}^{\text{st}}_i$, and 2D correspondence $\hat{p}^{\text{st}}_j$ for $p_i$ can be obtained by volume rendering similar to Eq.(\ref{eq:alpha_comp}).

\subsubsection{3D volume of dynamic objects.}
For dynamic objects, there are often appearance changes and deformations in video. Simple color features can no longer guarantee temporal consistency, and are also very susceptible to similar distractors. Image features can represent objects more discriminatively and consistently \cite{wang2019learning,jabri2020space}, which would also be beneficial in obtaining continuous and consistent motion trajectories. Therefore, we additionally encode the 3D feature $f^{\text{dy}}$ in $G_{\text{dy}}$ via $F_\theta^{\text{dy}}$. Through volume rendering, apart from the pixel color $\hat{C}^{\text{dy}}_i$ and the 2D correspondence $\hat{p}^{\text{dy}}_j$ for $p_i$, the 2D feature $\hat{F}^{\text{dy}}_i$ can also be obtained:

\begin{equation}
  \hat{F}^{\text{dy}}_i(p_i)=\sum_{k=1}^K T^{\text{dy}}_k \alpha(\sigma^{\text{dy}}_k) f^{\text{dy}}_k, \text { where } T^{\text{dy}}_k=\prod_{l=1}^{k-1}\left(1-\alpha(\sigma^{\text{dy}}_l)\right).
  \label{eq:feat}
\end{equation}

\subsubsection{Transformations.}
We also carefully design the static transformation $\mathcal{T}^{\text{st}}$ and the dynamic transformation $\mathcal{T}^{\text{dy}}$, respectively. For the static scenes, the movement caused by the camera's rigid-motion is mainly considered, so we model the static mapping by optimizing the parameters of the 3D affine transformation between frames, i.e., $\mathcal{T}^{\text{st}}(\cdot) = A_\theta(\cdot)$. The motion trajectories of the object are the result of the superposition of the camera and the object's own motion, which often corresponds to a more complex non-linear transformation. Therefore, we refer to the method of \cite{wang2023omnimotion}, and additionally utilize the Real-NVP\cite{dinh2016density} to model the dynamic mapping, i.e., $\mathcal{T}^{\text{dy}}(\cdot) = M_\theta(A_\theta(\cdot); \phi_i)$, where $\phi_i$ stands for the individually optimized latent code for each frame. Due to the ambiguity of motion decomposition, we find that for static transformation, it is also beneficial to appropriately add some non-linear transformation layers, which will be discussed in the subsequent experimental analysis.

\subsubsection{Volume fusion.}
After obtaining the static and dynamic representations, we obtain the final 3D volume $G_{\mathrm{cb}}$ by volume fusion, i.e., $\mathcal{M}(G_{\text{st}}, G_{\text{dy}})=G_{\mathrm{cb}}$, in order to get the final representation of the whole video scene, as well as the pixels' motion trajectories. We leverage the orthogonal approximation technique described in previous work \cite{martin2021nerf} to approximate the process:

\begin{equation}
  \sigma_k^{\mathrm{cb}} x_j^k = \beta^{\text{st}}_k u_j^k \sigma_k^{\text{st}}+(1-\beta_k^{\text{st}}) v_j^k \sigma_k^{\text{dy}},
  \end{equation}
where $u_j^k$/$v_j^k$ represents the local 3D points at $L_j$ obtained by static/dynamic transformation for $x_i^k$. The motion rendering in Eq.(\ref{eq:alpha_comp}) can be rewritten as:

\begin{equation}
  \begin{aligned}
      \hat{\boldsymbol{x}}_j^{\mathrm{cb}} & =\sum_{k=1}^K T_k^{\mathrm{cb}}\left(\beta_k^{\text{st}} \alpha\left(\sigma_k^{\text{st}}\right) u_j^k+\left(1-\beta_k^{\text{st}}\right)  \alpha\left(\sigma_k^{\text{dy}}\right)
      v_j^k\right) \\
  \text { where } T_k^{\mathrm{cb}}& =\exp \left(-\sum_{l=1}^{k-1} \beta_l^{\text{st}} \sigma_l^{\text{st}} +\left(1-\beta_l^{\text{st}}\right)  \sigma_l^{\text{dy}}\right)
  \end{aligned}
  \label{eq:ax}
  \end{equation}

We obtain final 2D correspondence $ \hat{\boldsymbol{x}}_j^{\mathrm{cb}} $ by projecting $ \hat{p}^{cb}_j$ to the image plane. Similarly, we can get the rendered color $\hat{C}^{\mathrm{cb}}_i$ by replacing 3D coordinate with color in Eq.(\ref{eq:ax}).

\subsection{Optimization}

In order to obtain a decoupled 3D representation of the video, we utilize a variety of loss functions, including feature rendering loss, color rendering loss, motion rendering loss, etc. This section will describe them in detail.

\subsubsection{Data preprocessing.}
We refer to the previous method \cite{wang2023omnimotion,wang2023flow} to get the optical flow between every two frames of the video by the advanced optical flow estimation method RAFT~\cite{teed2020raft} and filter out the optical flow results with high confidence as the training pseudo-labels. In addition to this, to realize feature rendering loss, we also extract the feature $F$ by video correspondence learning methods \cite{wang2019learning,xu2021rethinking,caron2021emerging}. In order to achieve better decoupling, we also refer to the previous methods \cite{li2023dynibar,gao2021dynamic} to get the mask $M$ of the moving objects by semantic segmentation ~\cite{chen2017deeplab,strudel2021segmenter} or motion segmentation \cite{ xie2022segmenting,yuan2023isomer,li2021motion}, which serves as a guidance for the network in the initial period of training.

\subsubsection{Optimization for dynamic objects.}

For dynamic objects, the following loss functions are used for optimization. After obtaining the collected reliable optical flow $\boldsymbol{f}_{i \rightarrow j}=p_j - p_i$, predict optical flow $\hat{\boldsymbol{f}}^{\text{dy}}_{i \rightarrow j}=\hat{p}^{\text{dy}}_j - p_i$ and the motion segmentation mask $M_i$, the $L_1$ distance is used as the motion rendering loss, and the motion segmentation mask forces the transformation network to focus only on modeling the motion of the dynamic objects during the initial phase of optimization, and the $\Omega_f$ indicates the set of filtered flows:

\begin{equation}
  \mathcal{L}^{\text{dy}}_{\text {flo }}=\sum_{\boldsymbol{f}_{i \rightarrow j} \in \Omega_f} M_i(p_i) \left\|\hat{\boldsymbol{f}}^{\text{dy}}_{i \rightarrow j}-\boldsymbol{f}_{i \rightarrow j}\right\|_1
\end{equation}

In addition to this, we also utilize the color rendering loss. The loss decreases only if corresponding 2D points $p_i$ and $p_j$ map to 3D points in the canonical space with the same color. We believe through this way, the representation implicitly captures the correspondences by matching the appearance, which shares similar spirits to previous matching-based motion estimation methods \cite{vondrick2018tracking,sun2018pwc}. The loss is defined as:

\begin{equation}
  \mathcal{L}^{\text{dy}}_{\text {pho}}=\sum_{p_i \in \Omega_p} M_i(p_i) \left\|\hat{C}_i^{\text{dy}}(p_i)-C_i(p_i)\right\|_2^2
\end{equation}
The $\Omega_p$ indicates the set of filtered points. However, as introduced above, the photometric error reflects only the similarity in color space and is very susceptible to distractors. For example, in Figure \ref{fig:feat_render}(a), the pixels $\mathbf{s_1}, \mathbf{s_2}, \mathbf{s_3}$ are very similar in color space, leading to a reduction of the photometric error even if $\mathbf{s_1}$ in the right frame is incorrectly mapped to $\mathbf{s_3}$. Whereas in (b), $\mathbf{s_3}$ is significantly different from the rest of the points in the feature space, which indicates this problem can be mitigated by rendering discriminative features. Moreover, inter-frame appearance changes and deformations make pixels inconsistent between frames, especially on dynamic objects. Therefore, this paper proposes a feature-based rendering loss:

\begin{figure}[!t]
  \centering
  \includegraphics[width=0.9\textwidth]{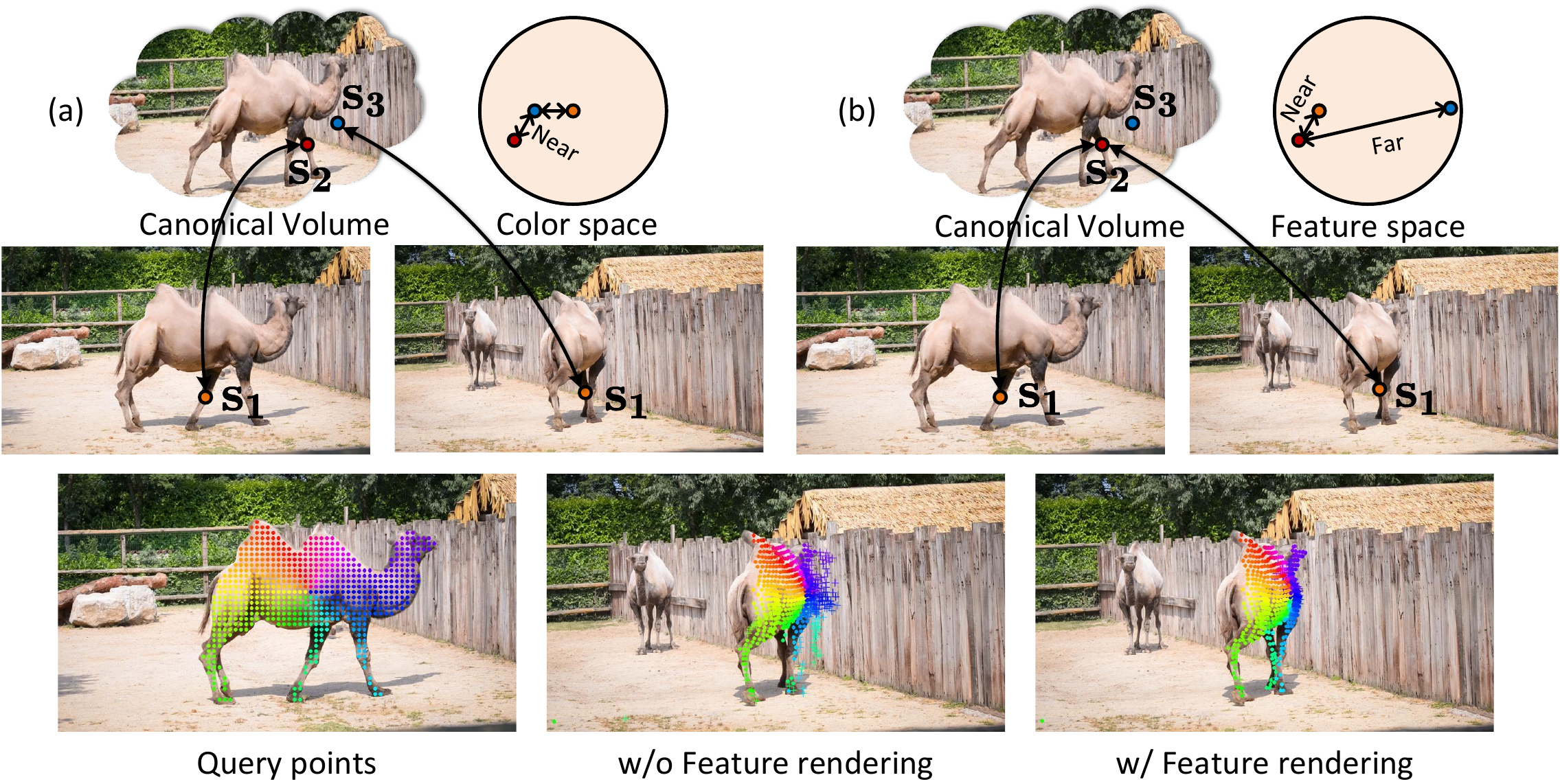}
  \caption{The illustration of feature rendering loss. We also show the qualitative results in the last row where the query points are marked with different colors. Please refer to the corresponding video in the supplemental materials. \textbf{(Zoom in for best view)}}
  \label{fig:feat_render}
\end{figure}

\begin{equation}
  \mathcal{L}_{\text {feat }}= \sum_{p_i \in \Omega_p} M_i(p_i) \left\|\hat{F}_i^{\text{dy}}(p_i)-F_i(p_i)\right\|_1
\end{equation}
Using the pre-trained temporally-consistent and spatially-discriminative features~\cite{caron2021emerging} as labels, we rectify the dynamic transformation and force corresponding pixels to map to the same 3D point in canonical space. The loss function used for dynamic objects is defined as $\mathcal{L}^{\text {dy}} = \mathcal{L}^{\text {dy}}_{\text {flow}} + \lambda_1^{\text{dy}} \mathcal{L}^{\text {dy}}_{\text {pho}} + \lambda_2^{\text{dy}} \mathcal{L}_{\text {feat}}$.

\subsubsection{Optimization for static scenes.}
For static scenes, we also adopt motion and color rendering loss for optimization:

\begin{equation}
  \mathcal{L}^{\text{st}}_{\text {flo }}=\sum_{\boldsymbol{f}_{i \rightarrow j} \in \Omega_f} (1-M_i(p_i)) \left\|\hat{\boldsymbol{f}}^{\text{st}}_{i \rightarrow j}-\boldsymbol{f}_{i \rightarrow j}\right\|_1
\end{equation}

\begin{equation}
  \mathcal{L}^{\text{st}}_{\text {pho}}=\sum_{p_i \in \Omega_p} (1-M_i(p_i)) \left\|\hat{C}_i^{\text{st}}(p_i)-C_i(p_i)\right\|_2^2
\end{equation}

Besides, the motion segmentation masks are utilized to supervise the rendered motion mask $\hat{B}_i^{\text{st}}(p_i)$, giving guidance to static network in the early stages:

\begin{equation}
    \mathcal{L}_{\text {seg}}= \sum_{p_i \in \Omega_p} -(1-M_i(p_i))log(\hat{B}_i^{\text{st}}(p_i)) - M_i(p_i)log(1-\hat{B}_i^{\text{st}}(p_i)),
\end{equation}
the optimized motion confidence will be further used in volume fusion, and the loss function used for static scenes is defined as $\mathcal{L}^{\text {st}} = \mathcal{L}^{\text {st}}_{\text {flow}} + \lambda_1^{\text{st}} \mathcal{L}^{\text {st}}_{\text {pho}} + \lambda_2^{\text{st}} \mathcal{L}_{\text {seg}}$.

\subsubsection{Optimization for final representation.}
Based on obtaining the respective representation and transformation of the static scenes and dynamic objects, we attempt to obtain the final representation by volume fusion.  We use the motion and color rendering loss for optimization, denoted as $\mathcal{L}^{\mathrm{cb}} = \mathcal{L}^{ \mathrm{cb}}_{\text {flo }} + \lambda_1^{\text{cb}} \mathcal{L}^{\mathrm{cb}}_{\text {pho }}$:

\begin{equation}
  \mathcal{L}^{\mathrm{cb}}_{\text {flo }}=\sum_{\boldsymbol{f}_{i \rightarrow j} \in \Omega_f}  \left\|\hat{\boldsymbol{f}}^{cb}_{i \rightarrow j}-\boldsymbol{f}_{i \rightarrow j}\right\|_1
\end{equation}

\begin{equation}
  \mathcal{L}^{\mathrm{cb}}_{\text {pho }}= \sum_{p_i \in \Omega_p} \left\|\hat{C}_i^{cb}(p_i)-C_i(p_i)\right\|_2^2
\end{equation}

In addition to the above losses, we also adopt regularization schemes used in prior work, e.g., the scene flow smoothness loss\cite{li2021neural}, the distortion loss \cite{barron2022mip},  which are labeled as $\mathcal{L}^{\text{reg}}$. The final overall training loss is defined as follows:

\begin{equation}
    \mathcal{L}=\mathcal{L}^{\text{dy}} + \mathcal{L}^{\text{st}} + \mathcal{L}^{\mathrm{cb}} + \mathcal{L}^{\text{reg}}
\end{equation}

\section{Experiment}
To validate the proposed method, we evaluate our method on the TAP-Vid benchmark \cite{doersch2022tap}, which is designed to evaluate the performance of long-range motion estimation (point tracking) in the real world. This section first describes the experimental setup including dataset, network, training, and evaluation details. After that, detailed ablation studies and performance comparisons with the baseline methods will be presented to prove the effectiveness of the proposed method for motion estimation.
\subsection{Implementation details}
\subsubsection{Dataset.}
Since the proposed method is a test-time optimization method, considering the tractable, a more representative TAP-Vid-DAVIS~\cite{doersch2022tap} dataset is selected for training and testing. This dataset, contains 30 videos in the real world, with videos ranging from 34-104 frames and an average of 21.7 points annotations per video. The videos in TAP-Vid \cite{doersch2022tap} are downsampled to 256 $\times$ 256, which ensures that no more information than expected is used.

\subsubsection{Network.}
The parameters corresponding to the affine transformation $A_\theta$ are predicted by the 2-layer MLP. Meanwhile, as discussed in Figure \ref{fig:ab3}, we also incorporate non-linear invertible transformation layers \cite{dinh2016density} in the static transformation network and set the number of layers $n_{\text{st}}$ to 3. The $M_{\theta}$ in the dynamic transformation network consists of 6 invertible layers. The $F^{\text{dy}}_{\theta}$ and $F^{\text{st}}_{\theta}$ use the GaborNet \cite{fathony2020multiplicative} with 3 layers and 512 channels per layer to encode the color, motion confidence, and density. Considering the complexity of the features, we leverage another GaborNet alone to do the feature rendering.  Following OmniMotion~\cite{wang2023omnimotion}, a 2-layer fully-connected layer with 256 channels is included to compute the implicit representation $\phi_i$ for each video frame, and the normalized time $t_i$ is used as the input to obtain the time embedding.

\subsubsection{Training details.}
DecoMotion uses the Adam as an optimizer to train each video sequence for 100k iterations. Warm-up strategy is used to slowly increase the learning rate, and we decay the learning rate every 20k steps thereafter. In each training batch, we extract 256 pairs of optical flows from 8 pairs of images. For each pixel corresponding to a ray, we uniformly sample $K=32$ local 3D points in the $z$ axis. In feature rendering, we use DINO~\cite{caron2021emerging} as an image feature extractor in the data preprocessing stage to extract 768-d features for each point. The rest of the settings including reliable sample filtering, hard sample mining strategy, and far and near depth for volume rendering refer to the work in \cite{wang2023omnimotion}. Further implementation details are presented in the supplemental materials.

\subsubsection{Evaluation metrics.} Following the TAP-Vid benchmark \cite{doersch2022tap} and OmniMotion \cite{wang2023omnimotion}, we report the tracking position accuracy ($ < \delta^x_{avg}$), Occlusion Accuracy (OA), Average Jaccard (AJ), and Temporal Coherence (TC) for predict motion trajectories. Please refer to the benchmark for more information.

\subsection{Ablation study}
We perform detailed ablation studies on point tracking. Considering DecoMotion is a test-time optimization method, to ensure the executability of the experiments, a subset of TAP-Vid-DAVIS~\cite{doersch2022tap} (480p) is randomly sampled.

\subsubsection{The effectiveness of decomposition.} 

\begin{table}[!t]
  \centering
  \small
  \caption{Ablation study for decomposition. }
  \resizebox{0.65\textwidth}{!}{
      \setlength\tabcolsep{9pt}
      \renewcommand\arraystretch{1.20}
  \begin{tabular}{ccc|ccc}
    \thickhline
  \multirow{2}{*}{$\mathcal{L}^{\text{st}}$} & \multirow{2}{*}{$\mathcal{L}^{\text{dy}}$} & \multirow{2}{*}{$\mathcal{L}^{\text{cb}}$} & \multicolumn{3}{c}{TAP-Vid-DAVIS} \\ 
        &        &         &      AJ $\uparrow$   &     $<\delta_{a v g}^x \uparrow$      &      OA  $\uparrow$   \\ \hline
             &                       &            &     52.6      &  69.9     &     85.3      \\
             $\checkmark$                     &            &        &     27.6  &    31.0    &     77.6      \\
                     & $\checkmark$         &                       &   54.3        &     73.3      &       85.2    \\

                     &         &    $\checkmark$    &   57.9        &     76.4      &       86.0    \\
                  
             $\checkmark$          & $\checkmark$         & $\checkmark$          &                 59.9      &     79.0      &     86.1  
            \\ \thickhline
  \end{tabular}%
  }
  \label{tab:ab_deco}
  \end{table}

We first study how each design in our decoupled representation affects the performance of point tracking. The original OmniMotion is set as the baseline. The results are shown in Table \ref{tab:ab_deco}. The $\mathcal{L}^{\text{st}}$, $\mathcal{L}^{\text{dy}}$ and $\mathcal{L}^{\text{cb}}$ represents the loss functions used in each representation optimization. Solely optimizing representation for static scenes substantially harms the performance, the learned static transformation can not deal with the complex motion patterns of dynamic objects. Whereas the proprietary transformation for dynamic objects improves the tracking performance from 69.9\% to 73.3\%. With the help of the motion segmentation mask, the dynamic transformation can focus on learning the difficult parts, achieving the globally optimal capability of estimating the object's motion. However, the dynamic transformation still fails to capture the camera motion. By combining the static part with $\mathcal{L}^{\text{cb}}$ in volume fusion, the tracking accuracy is further improved to 79.0\%. The results demonstrate the effectiveness of the divide-and-conquer strategy in motion estimation.

\begin{figure}[!t]
    \centering
    \includegraphics[width=1.0\textwidth]{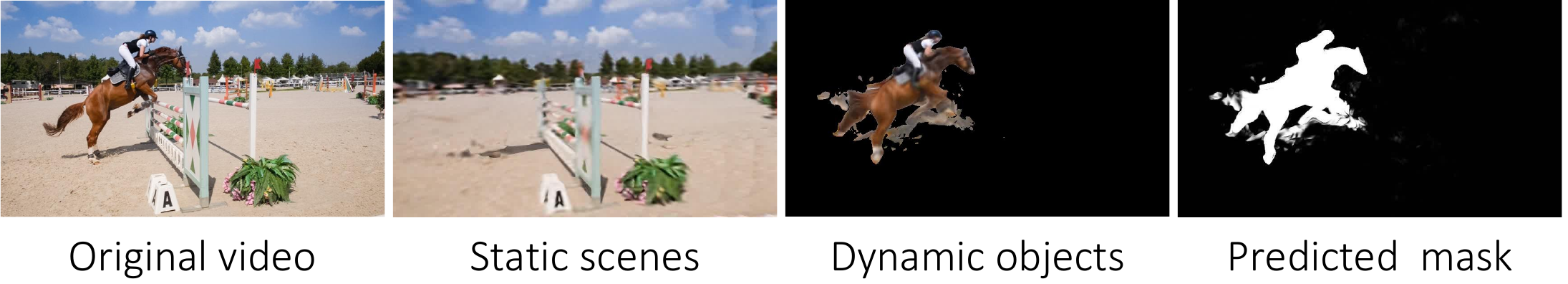}
    \caption{The rendering results for static scenes and dynamic objects. Only optimized with $\mathcal{L}^{\text{cb}}$, we still observe the decomposition in terms of motion and appearance. }
    \label{fig:ab1}
  \end{figure}

The motion segmentation models~\cite{yuan2023isomer} are complex and sometimes could be brittle. A dynamic/static point could be incorrectly assigned to a static/dynamic point, which misleads the optimization. Without the guidance of motion segmentation masks, we find optimizing the final representation solely with $\mathcal{L}^{\text{cb}}$ still shows competitive results. In Figure \ref{fig:ab1}, we show rendering results of static appearance $\hat{C}^{\text{st}}_i$, dynamic appearance $\hat{C}^{\text{dy}}_i$, and motion segmentation mask $\hat{B}^{\text{st}}_i$, we can still observe the decomposition in terms of appearance and motion. Moreover, thanks to the insightful feedback of reviewers, we are seeking more elegant ways to provide reasonable results without a segmenter, e.g. first optimizing the globally-rigid component alone, and then adding the non-rigid component to explain the remainder. We regard it as our future work.

  \begin{table}[!b]
    \caption{Ablation study for feature rendering.} 
  
    \label{tab:feat_render}
    \begin{subtable}{.5\linewidth}
      \centering
        \caption{Abalation for feature rendering in different scenes.}
        \resizebox{!}{1.52cm}{
          \setlength\tabcolsep{4pt}
          \renewcommand\arraystretch{1.50}
      \begin{tabular}{cc|cc|ccc}
  \thickhline
      \multicolumn{2}{c}{$\mathcal{L}_{\text{pho}}$} & \multicolumn{2}{c|}{$\mathcal{L}_{\text{feat}}$} & \multicolumn{3}{c}{TAP-Vid-DAVIS} \\ \hline
            Dynamic & Static        & Dynamic     & Static         &      AJ $\uparrow$   &     $<\delta_{a v g}^x \uparrow$      &      OA  $\uparrow$   \\ \hline
                 &           &            &            &     51.2      &  69.7     &     85.6      \\
                 $\checkmark$          &           &            &        &     55.6   &    74.5       &     85.8      \\
                 $\checkmark$          & $\checkmark$         &            &            &   57.3        &     76.1      &       85.5    \\
                 $\checkmark$          & $\checkmark$         & $\checkmark$          &            &     59.9      &     79.0      &     86.1     \\
                 $\checkmark$          & $\checkmark$         & $\checkmark$          & $\checkmark$          &   60.2        &    79.2       &      85.9     \\ \thickhline
      \end{tabular}%
        }
    \end{subtable}%
    \begin{subtable}{.5\linewidth}
      \centering
        \caption{Ablation for different features.}
        \resizebox{!}{0.027cm}{
          \setlength\tabcolsep{20pt}
          \renewcommand\arraystretch{1.40}
           \begin{tabular}[t]{c|c}
            \thickhline
                  Method &  $<\delta_{a v g}^x \uparrow$   \\
                  \hline
                  w/o $\mathcal{L}_{\text{feat}}$  & 76.1 \\
                  $\mathcal{L}_{\text{feat}}$ w/ FGVC \cite{li2023learning}      & 76.7 \\
                  $\mathcal{L}_{\text{feat}}$ w/ VFS \cite{xu2021rethinking}    & 77.3 \\
                  $\mathcal{L}_{\text{feat}}$ w/ DINO \cite{caron2021emerging}    & 79.0 \\
                  \thickhline
           \end{tabular}
        }
    \end{subtable} 
  \end{table}

\subsubsection{Feature rendering.} We analyze the effect of the proposed feature rendering. The results are shown in Table \ref{tab:feat_render}.  In (a), we regard the DecoMotion without using any appearance rendering loss as the baseline. With the implicit appearance matching in canonical space, both color and feature rendering enhance the performance of motion estimation. Moreover, the proposed feature rendering further boots up the performance from 76.1\% to 79.0\%. As shown in the last row of 
Figure \ref{fig:feat_render}, the given query points are misled by the distractors in the background. By using feature rendering loss, the query points are accurately matched to the object in another frame, even facing large deformations. Nevertheless, the feature rendering on static scenes does not help much. We believe this is because of the relatively poor representation ability of the pre-trained features \cite{caron2021emerging} and the slight inter-frame changes in the background. Besides, in (b), we use different pre-trained features as supervision, compared with FGVC~\cite{li2023learning} that more specializes in extracting local features, the VFS~\cite{xu2021rethinking} and DINO~\cite{caron2021emerging}, which are better at representing objects, show better performance.

\subsubsection{The design of transformation.}
\begin{figure}[!t]
  \centering
  \includegraphics[width=0.8\textwidth]{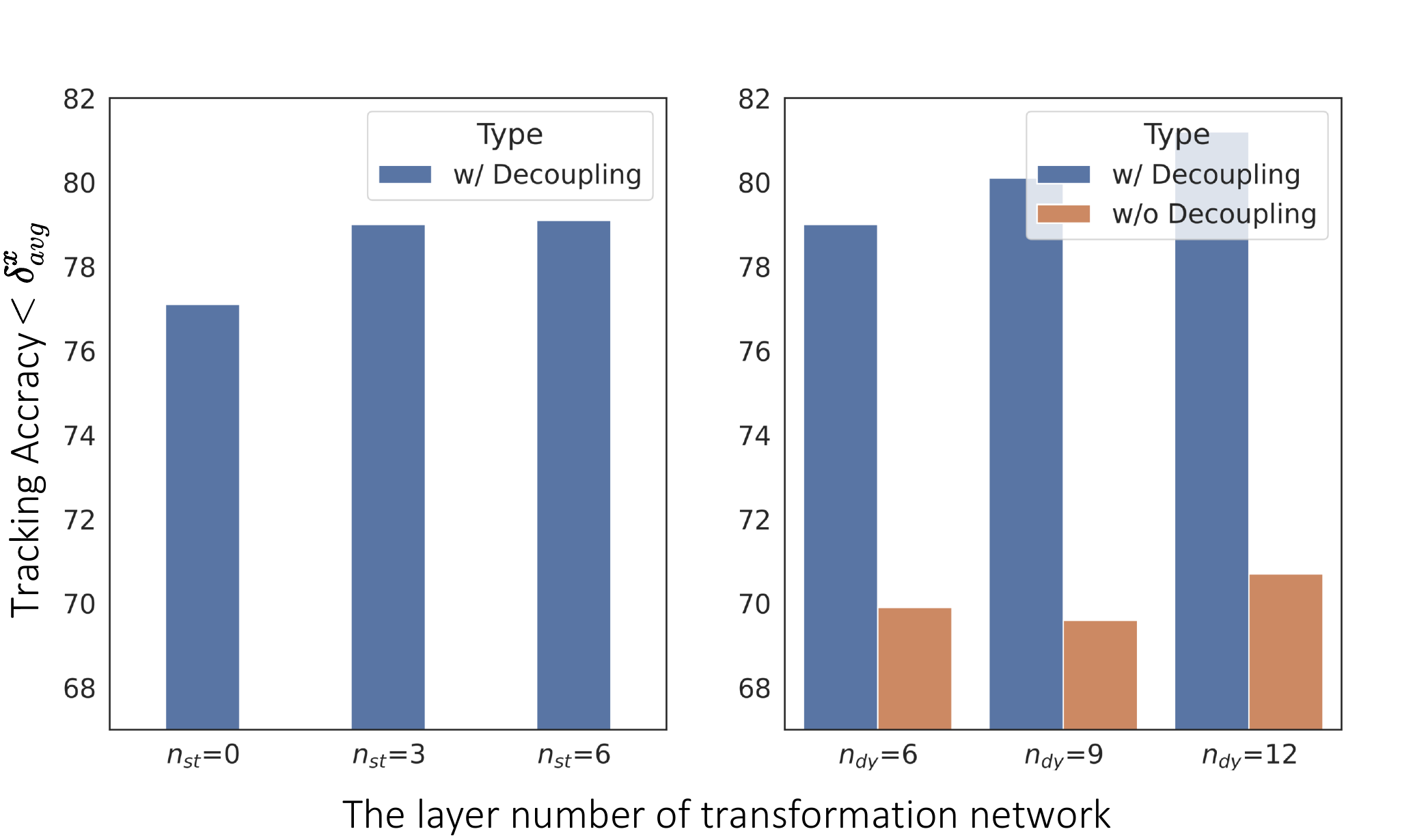}
  \caption{The ablation study for the transformations. The $n_{\text{st}},n_{\text{dy}}$ represent the number of non-linear layers \cite{dinh2016density} used in static and dynamic transformations. }
  \label{fig:ab3}
\end{figure}

 In Figure \ref{fig:ab3}, the $n_{\text{st}},n_{\text{dy}}$ represent the number of non-linear invertible layers \cite{dinh2016density} used in static and dynamic transformations. For static transformation, purely utilizing affine transformation, i.e., $n_{\text{st}}=0$, gives the worst performance. We find the motion confidence prediction is not always perfect. If a dynamic point is incorrectly predicted as a static point, the affine transformation alone can not deal with it well. Adding some non-linear layers in the static transformation improves the performance. However, including more layers does not bring benefits. Besides, increasing the $n_{\text{dy}}$ in dynamic transformation shows a more significant improvement compared with the method without decoupling.

\subsection{Comparisons on point tracking}

\begin{table}[!t]
  \centering  \caption{Quantitative results for point tracking on TAP-Vid \cite{doersch2022tap}. The ``(RAFT)'' indicates using pairwise correspondences from RAFT \cite{teed2020raft} as input or supervision.}
  \resizebox{0.9\textwidth}{!}{
      \setlength\tabcolsep{5pt}
      \renewcommand\arraystretch{1.30}
  \begin{tabular}{lcccc|cccc}
  \thickhline
  \multirow{2}{*}{Method} & \multicolumn{4}{c}{DAVIS Strided} & \multicolumn{4}{c}{DAVIS First}   \\ \cline{2-9} 
  & AJ  $\uparrow$   & $<\delta_{a v g}^x \uparrow$       & OA  $\uparrow$   & TC  $\downarrow$   & AJ $\uparrow$      & $<\delta_{a v g}^x \uparrow$   & OA $\uparrow$  & TC  $\downarrow$   \\ \hline
  Flow-Walk-C \cite{bian2022learning}         & 35.2       & 51.4       & 80.6      &    \textbf{0.90}    &      -      &        -              &  -   &         -             \\
  Flow-Walk-D \cite{bian2022learning}        & 24.4       & 40.9       & 76.5      & 10.41 &- & - &  -   & -   \\
  RAFT-C \cite{teed2020raft}             & 30.7       & 46.6       & 80.2      &  0.93      &27.1 &    42.1         &  77.6    &  \textbf{0.95}                    \\
  RAFT-D \cite{teed2020raft}              & 34.1       & 48.9       & 76.1      &  9.83     &     31.4      &     44.5             &   74.1     &      10.3             \\
  TAP-Net \cite{doersch2022tap}           & 38.4       & 53.4       & 81.4      &    10.82    &  33.0       &    48.6         &  78.8      &     -             \\

    PIPs \cite{harley2022particle}               & 39.9       & 56.0     & 81.3      &   1.78   &     36.4         &     53.8                 &     76.1   &     -             \\

  Context-TAP \cite{bian2023context}               & 48.9      & 64.0     & -     &     -   &     42.7       &   60.3          &    -    &        -          \\

        FGVC \cite{li2023learning}            & -       & 66.4      & -     &    -    &  -     &    62.8    &   -     &     -             \\


  PIPs$++$ \cite{zheng2023pointodyssey}               & -      & \textbf{73.7}     & -     &     -   &     -       &   69.1          &    -    &        -          \\
  TAPIR \cite{doersch2023tapir}              & \textbf{61.3}      & 73.6   & \textbf{88.8}     &     -   &     \textbf{56.2}       &   \textbf{70.0}          &    \textbf{86.5}    &        -          \\

  \hline
    Deformable-Sprites \cite{ye2022deformable}  & 20.6       & 32.9       & 69.7      & 2.07 &- & - &  -  &  -    \\ 
  OmniMotion \cite{wang2023omnimotion} (RAFT)   & 51.7       & 67.5       & 85.3   &  0.74 & 44.9 & 62.7  & 80.7   & 0.73    \\
  MFT \cite{neoral2024mft}  (RAFT)           & 56.1      & 70.8      & 86.9     &   -    &  47.3     &    66.8    &     77.8   &     -             \\
  DecoMotion (RAFT) & \textbf{60.2} & \textbf{74.4} & \textbf{87.2} & \textbf{0.69} & \textbf{53.0} & \textbf{69.9} & \textbf{84.2} & \textbf{0.69} 
    \\
  \thickhline
  \end{tabular}%
  }
  \label{tab:sota}
  \end{table}

  \begin{figure}[!t]
    \centering
    \includegraphics[width=1.0\textwidth]{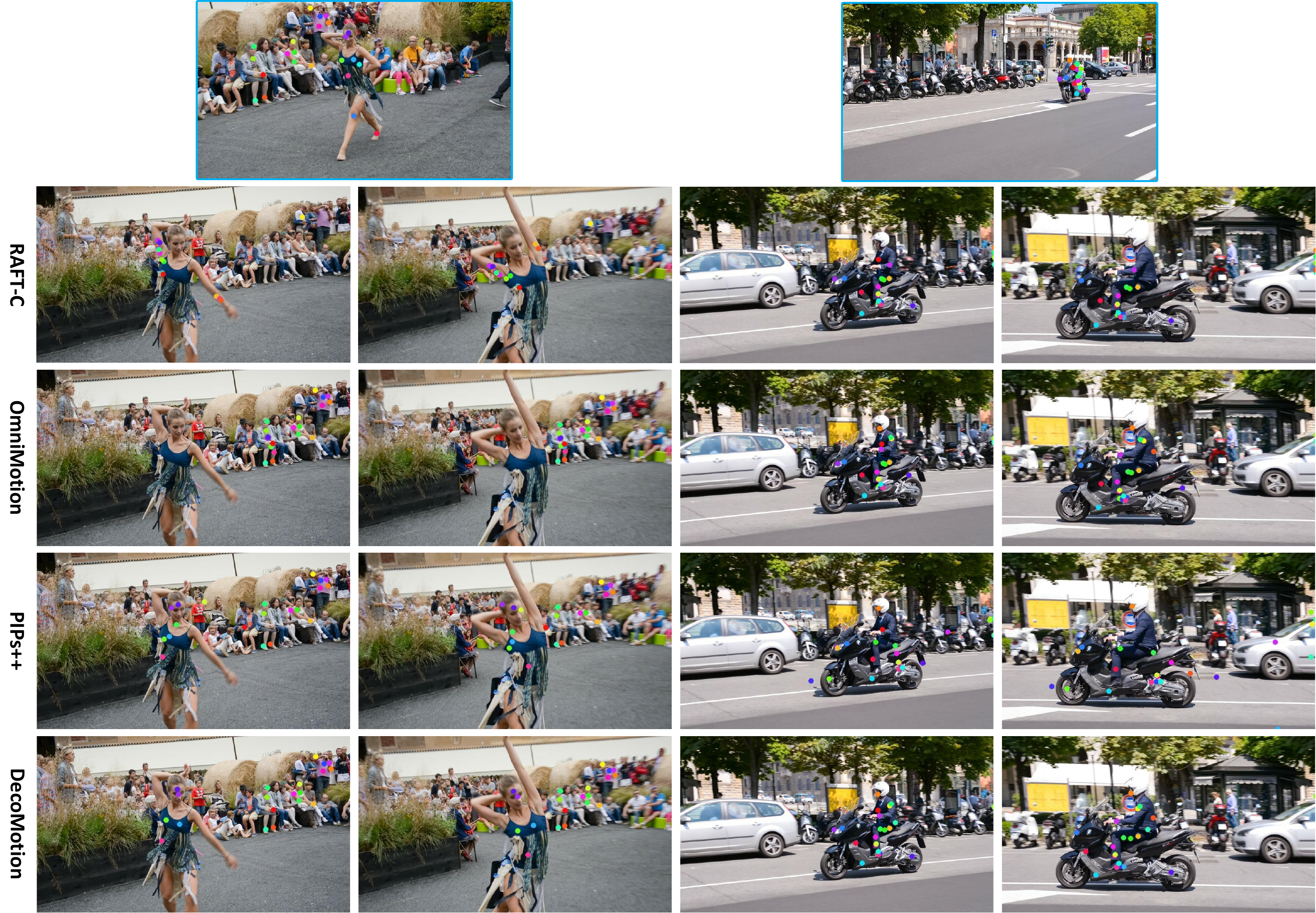}
    \caption{Qualitative results for point tracking. Given the query points marked with different colors in the first frame (blue border), we visualize the visible correspondences in randomly sampled frames. Please refer to more videos in the supplemental materials.  \textbf{(Zoom in for best view)}}
    \label{fig:quan1}
  \end{figure}

We compare our method with the state-of-the-art methods of point tracking. The quantitative results are shown in Table \ref{tab:sota}.  Apart from optimization-based methods Deformable-Sprites and OmniMotion, we add MFT that utilizes input pairwise correspondences from RAFT as our competitors. Compared with the most direct competitor OmniMotion, DecoMotion leads the tracking position accuracy $< \delta_{avg}^x$ by 6.9\%/7.2\% on DAVIS-Strided and DAVIS-First. Also, DecoMotion is superior to Deformable-Sprites and MFT, surpassing MFT by 3.6\%/3.1\%. We also include some dedicated point-tracking methods that track through occlusions by estimating multi-frame point trajectories for comparisons. Note DecoMotion is only supervised by RAFT which has limited capability of providing accurate point correspondences, DecoMotion still shows on-par performance, leading the tracking position accuracy by 0.7\% on DAVIS-Strided. 

We select several representative videos for inference. DecoMotion produces more accurate and stable pixel correspondences over frames. For example, in the left case of Figure \ref{fig:quan1}, the dancing girl presents a very complex movement with large deformations, the baseline methods either lose track or get imprecise results, while DecoMotion can consistently track points accurately, which validates the superior ability of DecoMotion to handle the motion of dynamic objects.

\subsection{Visualize the appearance decomposition}

\begin{figure}[!t]
  \centering
  \includegraphics[width=1.0\textwidth]{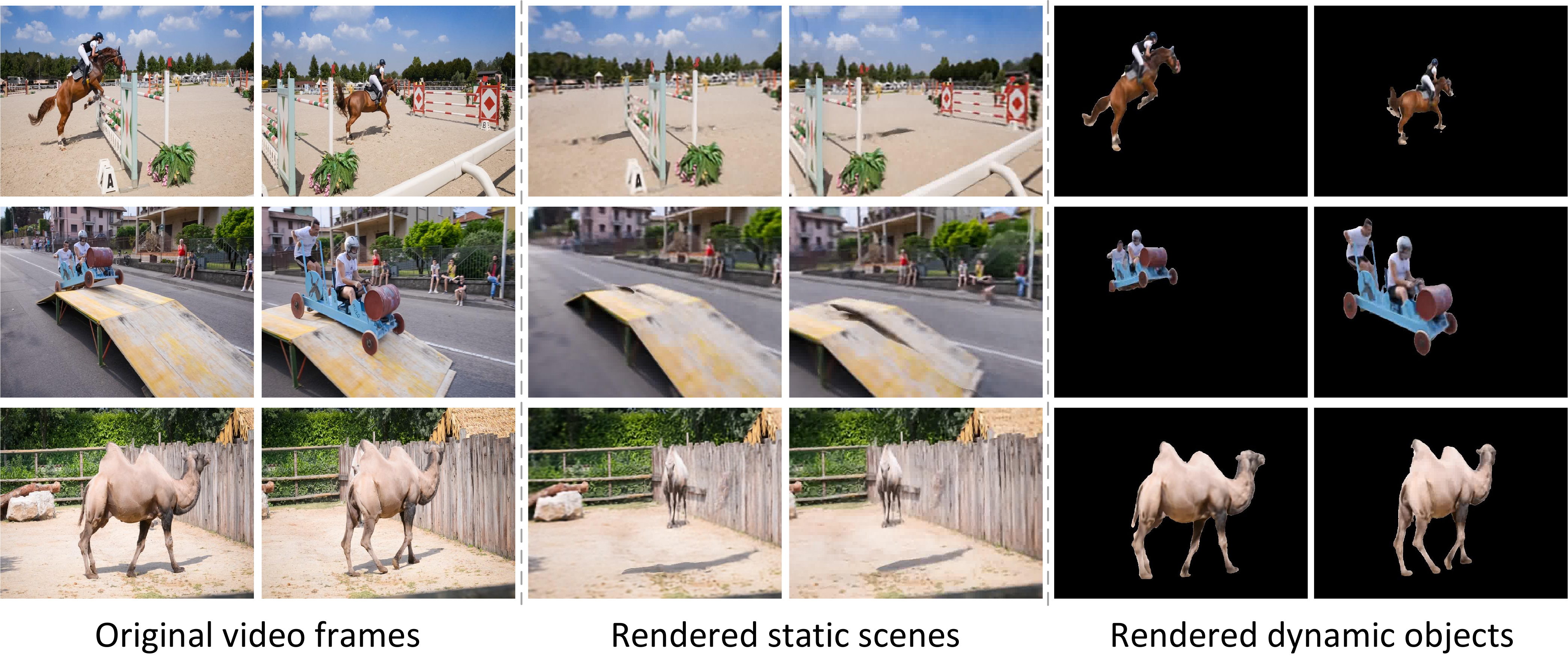}
  \caption{The rendering results for decoupled static scenes and dynamic objects. Please refer to the corresponding videos in the supplemental materials.}
  \label{fig:vis_deco1}
\end{figure}

We present the visualization of the decoupled static scenes and dynamic objects in Figure \ref{fig:vis_deco1}. Using the individual 3D canonical volume and transformation, we generate each frame in the given video by volume rendering. We can clearly observe the static part focuses on learning the appearance of static scenes, while the dynamic part pays more attention to the dynamic objects. With such a decoupled representation, our method can ``remove'' the dynamic objects while maintaining reasonable structure. 

\section{Conclusion}

In this work, we introduce an innovative test-time optimization approach for establishing accurate pixel correspondences. We separately establish 3D representation for static scenes and dynamic objects. For dynamic objects, we address their complex pattern by incorporating more non-linear layers with a feature rendering loss. For static scenes, we employ a simpler network to handle rigid motion, while also modeling the confidence of whether 3D points are static or not. Then we fuse these two canonical volumes to better represent motion and appearance. Quantitative results on point tracking and qualitative results of appearance decomposition validate the proposed method.

\noindent \textbf{Acknowledgements}. \small We thank anonymous reviewers for their valuable feedback. This work was supported by the Natural Science Foundation of China under Grant 61931014, and by the Fundamental Research Funds for the Central Universities under Grant WK3490000006. 

\section*{Supplementary Material}
\setcounter{section}{0}

The supplementary material contains: 1) more details about the training;  2) the demo video of feature rendering;
3) the demo video of appearance decomposition;
4) the demo video consists of more qualitative results of point tracking. 5) more results of tracking on multiple dynamic objects. 6) the limitations of DecoMotion.

\section{More training details}
We include more training details here. The base learning rates for the density/color network $F^{\text{st}}_{\theta},F^{\text{dy}}_{\theta}$, the transformation network $M_{\theta}$, and the time embedding MLP are set to $3 \times 10^{-4}, 1 \times 10^{-4}$, and $1 \times 10^{-3}$, respectively. For loss weight,  $\lambda_1^{\text{dy}}/\lambda_2^{\text{dy}}$ for dynamic objects are set to 1/10, $\lambda_1^{\text{st}}/\lambda_2^{\text{st}}$ for static scenes are set to 1/1, $\lambda_1^{\text{cb}}$ for final representation is set to 1. In the data-preprocessing stage, we use Isomer~\cite{yuan2023isomer} to obtain the motion segmentation mask. To properly leverage Isomer to process videos of TAP-Vid~\cite{doersch2022tap} benchmark for comparisons with state-of-the-art point trackers, we align with the initial training configuration of Isomer~\cite{yuan2023isomer} by upsampling the original video with a resolution of $256 \times 256$ to $384 \times 512$. Note we accomplish it by bilinear interpolation to ensure no more information than expected is used in data-preprocessing and training.

\section{Qualitative results of feature rendering}
We enclose the corresponding video for Figure 3 in the main paper to verify the effectiveness of feature rendering. We name it ``\textcolor{pink}{demo-feat.mp4}'' in our supplementary material. From the video, we can observe leveraging feature rendering loss exhibits clear improvements in tracking accuracy especially for the deformed object.


\section{Qualitative results of appearance decomposition}

We enclose the videos with appearance decomposition. We name it ``\textcolor{pink}{demo-appearance-deco.mp4}'' in our supplementary material.  In these videos, the fixed component primarily delves into understanding the visual attributes of stationary scenes, whereas the variable component directs its attention toward dynamic objects. By employing this decoupled representation, our approach achieves the capability to effectively ``eliminate'' dynamic objects while preserving a coherent structure. However, we also observed that for some of the more minor motions, our method does not decouple them very accurately, such as in the fourth sequence in the video, where the motion of the object (the camel lifting its head) is still observed in the rendered static scenes.

\section{Qualitative results of point tracking}
We enclose several representative video clips on TAP-Vid-DAVIS~\cite{doersch2022tap} to verify the effectiveness of our method compared with state-of-the-art methods, e.g., RAFT~\cite{teed2020raft}, OmniMotion~\cite{wang2023omnimotion} and PIPs$++$~\cite{zheng2023pointodyssey}. Please note all experiments are conducted on higher resolution (480p). We name it ``\textcolor{pink}{demo-point-tracking.mp4}'' in our supplementary material. As observed in the enclosed video, the results produced by our method show clear improvements over state-of-the-art methods. The predictions of our method tend to have more accurate and smooth trajectories. These examples again verify the effectiveness of our method. 

\section{Qualitative results of tracking on multiple dynamic objects}

\begin{figure}[!t]
    \centering
    \includegraphics[width=1.0\textwidth]{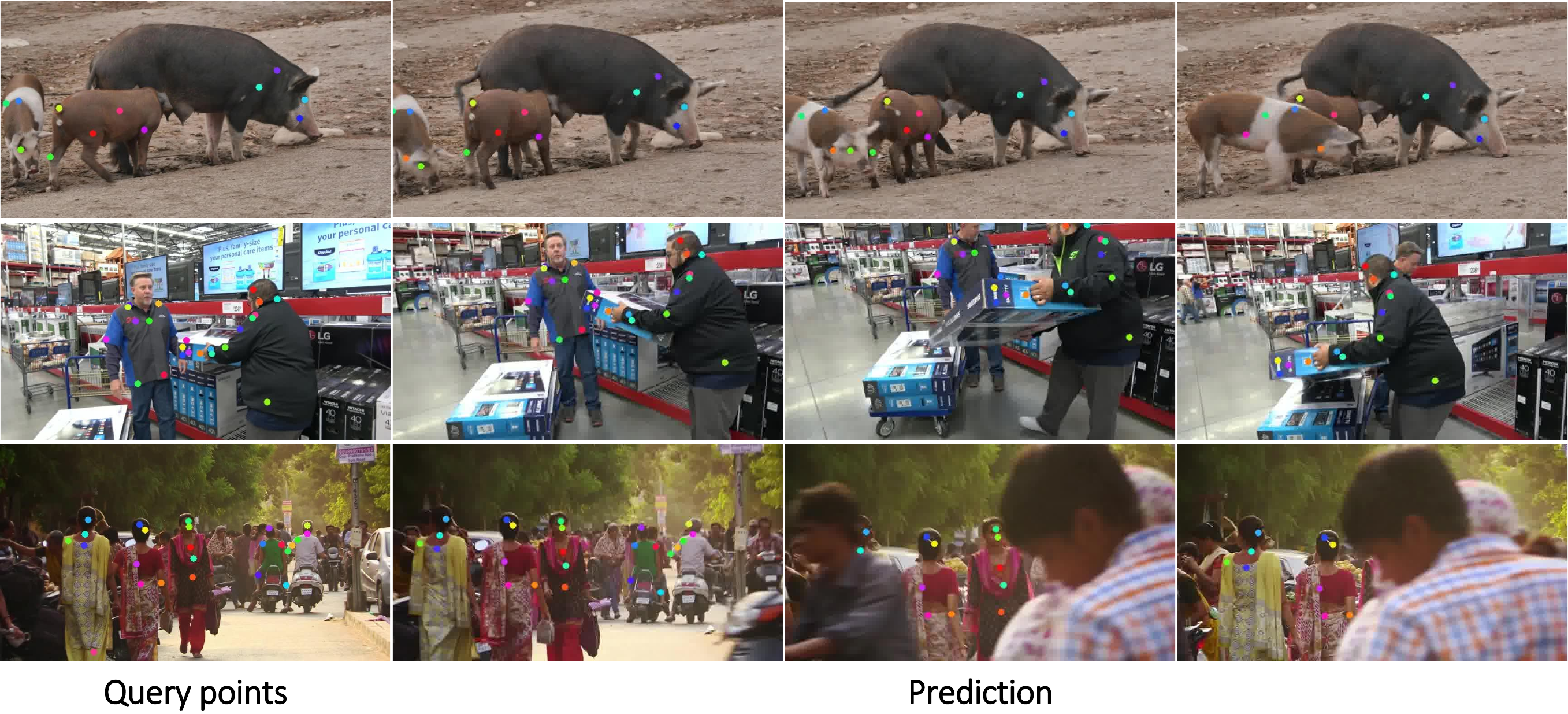}
    \caption{Qualitative results for point tracking on multiple dynamic objects. \textbf{(Zoom in for best view)} }
    \label{fig:multi}
  \end{figure}

  We show more results for point tracking on multi-dynamic objects. As shown in the Figure~\ref{fig:multi}, the method proposed in this paper is still able to effectively cope with the motion of multiple dynamic objects and obtain accurate point trajectories.

  \section{Limitations}
  While obtaining significant improvement, DecoMotion still highly relies on pre-computed optical flows. However, the recent pairwise optical flow methods \cite{teed2020raft,xu2022gmflow} struggle to model the rapid and large motion, failing to provide us with reliable and dense correspondences in some cases. Using the advanced multi-frame motion estimator \cite{doersch2023tapir,zheng2023pointodyssey} may alleviate this problem. In addition, we observe the training may fail or trap at local minima if there is a scene switch in the video. Besides, we also observe that DecoMotion improves less on the more challenging scenes with multiple objects, which is worthy of further study.

%
%
\bibliographystyle{splncs04}
\bibliography{main}
\end{document}